# Fast Coherent Point Drift

Xiang-Wei Feng[1,*], Da-Zheng Feng[1], Yun Zhu[2]

1, National Laboratory of Radar Signal Processing, Xidian University, Xi'an, China

2 School of Computer Science, Shaanxi Normal University, Xi'an, China

Corresponding author: Xiang-Wei Feng (e-mail: fengxiangwei@stu.xidian.edu.cn).

**Abstract:** Nonrigid point set registration is widely applied in the tasks of computer vision and pattern recognition. Coherent point drift (CPD) is a classical method for nonrigid point set registration. However, to solve spatial transformation functions, CPD has to compute inversion of a $M \times M$ matrix per iteration with time complexity $O(M^3)$. By introducing a simple corresponding constraint, we develop a fast implementation of CPD. The most advantage of our method is to avoid matrix-inverse operation. Before the iteration begins, our method requires to take eigenvalue decomposition of a $M \times M$ matrix once. After iteration begins, our method only needs to update a diagonal matrix with linear computational complexity, and perform matrix multiplication operation with time complexity approximately $O(M^2)$ in each iteration. Besides, our method can be further accelerated by the low-rank matrix approximation. Experimental results in 3D point cloud data show that our method can significantly reduce computation burden of the registration process, and keep comparable performance with CPD on accuracy.

**I. Introduction**

The nonrigid point registration is an important technique in the fields of computer vision and pattern recognition. It has been broadly applied in image retrieval, medical image processing, 3D reconstruction, face alignment, and so on. Suppose that there are two point sets that are denoted as the model point set $\mathbf{X} = \left(\mathbf{x}_1^T, \mathbf{x}_2^T, \ldots, \mathbf{x}_M^T\right)^T \in \mathbb{R}^{M \times D}$, and the scene point set $\mathbf{Y} = \left(\mathbf{y}_1^T, \mathbf{y}_2^T, \ldots, \mathbf{y}_N^T\right)^T \in \mathbb{R}^{N \times D}$, respectively. The aim of nonrigid point registration method is to recover the spatial transformation between the two point sets. Many methods were developed under the iterative scheme that contains two closely coupled substeps: establish the correspondences, and recover the spatial transformation.

The iterative closest point (ICP) algorithm [1-2] is the most popular method for point set registration. It is simple, general and efficient. Based on nearest neighbor strategy, ICP establishes a binary matrix to represent the corresponding relationship between point sets. However, ICP requires that the initial positions of the point sets are close enough. Otherwise, it easily gets stuck into a poor local minimum.

By replacing the nearest neighbor strategy into softassign, TPS-RPM (Thin plate spline - Robust point matching) was developed in [3-4]. Softassign relaxes the binary constraint of ICP, and uses continuous values to represent point-to-point corresponding relationship. This can improve the robustness of the registration methods by making optimization process be continuous, rather than jumping in binary variables with great arbitrariness. Another impressive work is coherent point drift (CPD) [5], and an improved version in [6]. Based on motion coherence theory (MCT) and Gaussian Mixture Models (GMM), CPD used maximum likelihood estimation (MLE) to model the alignment of two point sets, and employed an Expectation-Maximization (EM) scheme to solve the problem. Both TPS-RPM and CPD adopt the same cost function and a similar alternating update strategy [7]. The main difference between them is that TPS-RPM uses TPS as the transformation function, and CPD uses Gaussian radial basis functions (GRBF). The parameters of TPS can be explicitly decomposed into rigid and nonrigid parts. But TPS does not exist in 4D and higher dimensions, while GRBF can be easily extended into any dimension. By refining the point distribution models, non-uniform Gaussian mixture models [8], asymmetric Gaussian mixture models [9], and Student's-t mixture models [10], are introduced to substitute GMMs to achieve the point set registration.

Recently, many works were developed on the basis of TPS-RPM and CPD. The first strategy is to estimate the correspondences by fusing various structural features, and the second is to introduce spatially constraints to preserve local structural topology (LSD). The above two strategies can be utilized together. In [11]-[16], the local structural descriptors, such as shape context (SC) [17], the inner distance based context descriptor (IDSC) [18], Fast Point Feature Histograms (FPFH) [19], and the three-dimensional context descriptor (3DSC) [20], or self-designed shape descriptors are exploited as an auxiliary to search the corresponding relationship. The motivation of the second strategy is that the neighborhood structure of a point is generally well preserved because of physical constraints [11]. In [11], the authors interpreted the neighboring points as a simple graph, and preserved local structures by maximizing the number of matched edges. In [21], the authors used a weighted least square error item to represent the connections of k-connected neighboring points. In [22]-[25] and [14], the classical manifold regularization techniques, such locally linear embedding (LLE), local preserving projections (LPP), Laplacian eigenmaps (LE), were employed to maintain stability of local structures in registration process. Although these methods can utilize local structural features more efficiently, there are two problems: first, the computational complexity would increase; second, they are sensitive to outliers because outliers would seriously affect the descriptions of local structures. Besides, by recasting the two point sets as two continuous densities using GMM, [7] and [26]-[29] reported a kind of novel methods that accomplished the nonrigid registration by minimizing the discrepancy between them. Limited by the length of the article, many excellent works are not introduced in our paper. More other works can refer to the good reviews [30]-[31].

The previous works are constructive and notable. However, the computational burden of the nonrigid point registration is still too heavy in large-scale problems. Roughly, there are two kinds of spatial transformation, which are the rigid and nonrigid. The rigid transformation contains translation, rotation and scaling. It can be modeled with few parameters. Correspondingly, a small number of correspondences can well recover the rigid transformation. Compared with rigid transformation, nonrigid transformation has two properties: First, the nonrigid transformation is nonlinear and arbitrary. In order to well fit the complex nonrigid transformation, we have to adopt complicated interpolation functions. These functions usually possess a number of free parameters to be solved. Second, in different local regions, the nonrigid transformation is likely to vary dramatically. Thus, in order to well represent the subtle changes, a number of controlling points have to be extracted from the naive images. These two properties would increase the computational burden.

To address this issue, we develop a fast method to significantly improve the computational efficiency of CPD, which is named fast-CPD. The computational bottleneck for CPD is in solving the transformation function. It has to calculate the inversion of a $M \times M$ matrix with time complexity $O(M^3)$ per iteration. The computational burden is too heavy if there are a large number of feature points to be matched. Compared with CPD, we introduce a corresponding constraint to force each model point to search its corresponding point in the scene point set. Based on this constraint, we can obtain the solutions of transformation function by updating a diagonal matrix with linear computational complexity, and taking matrix multiplication operation with approximately $O(M^2)$. Although we need to take eigenvalue decomposition of a $M \times M$ matrix once before the iteration begins, the computational cost can be significantly saved because matrix-inverse operation can be avoided in each iteration. Besides, like CPD, the computational cost of our method can be further reduced by using low-rank matrix approximation.

The rest of the paper is organized as follows: In Section II, we introduce the models for nonrigid point set registration. Section III details the fast implementations, low-rank approximation, and the relations between CPD and our method. In Section IV, we present the experimental results. Section V concludes this paper.

**II. Models**

The point features, including edges [32], corners [33], SIFT [34], ORB [35] and so on, are extracted from the corresponding images. The point sets can be regarded as the simplified representations of the images because abundant structural information and geometric characteristics are preserved. In practice, the point sets can be utilized to recover the spatial transformation between two images, which can significantly save the computational cost, and avoid the effects of pixel intensity changes, such as the illumination changes. However, because of scene variations, pollutions of the naive images, and limitations of point feature extractors, the degradations, such as noise, outlier, and occlusion, inevitably exist in the point sets. They can seriously break the structures in some local areas, which would make it be difficult to obtain adequate reliable one-to-one correspondences. To address this issue, the probability-based methods are developed, and have achieved good performance in accuracy. CPD is one of the representative probability-based works. Here, we briefly introduce the theory and the optimization scheme of CPD.

Given two point sets (the model point set $\mathbf{X} \in \mathbb{R}^{M \times D}$, and the scene point set $\mathbf{Y} \in \mathbb{R}^{N \times D}$), the ultimate goal is to search the interpolation functions $f(\mathbf{X}; \boldsymbol{\theta})$ to recover the spatial transformation from $\mathbf{X}$ to $\mathbf{Y}$, where $\boldsymbol{\theta}$ denotes the parameters of the interpolation functions. For convenience, we use $\tilde{\mathbf{x}}_m = f(\mathbf{x}_m; \boldsymbol{\theta})$ to denote the new position of $\mathbf{x}_m$ after transformation. In the probability-based methods, the point set registration problems are recast to a probability density estimation problem based on Gaussian mixture models (GMM). The model points are served as the centroids of GMMs. The probability distribution of the scene points can be represented as a linear combination of GMMs [36], which is denoted as

$$p(\mathbf{y}_n) = \sum_{m=1}^{M} \pi_m g(\mathbf{y}_n; \tilde{\mathbf{x}}_m, \sigma^2 \mathbf{I}) + \pi_{M+1}, \tag{1}$$

where $g(\mathbf{y}_n; \tilde{\mathbf{x}}_m, \sigma^2 \mathbf{I}) = \frac{1}{(2\pi\sigma^2)^{D/2}} \exp\left(-\frac{\|\mathbf{y}_n - \tilde{\mathbf{x}}_m\|^2}{2\sigma^2}\right)$, and $\pi_m$ is the probability that $\mathbf{y}_n$ is assigned to the $m-th$ distribution. In CPD, equal probabilities are assigned between $\mathbf{y}_n$ and the GMM components, which is denoted as $\pi_m = 1/M$, where $m = 1,\ldots,M$. In order to deal with noise and outliers, an extra term is added, which is denoted as $\pi_{M+1} = 1/N$. An empirical parameter $\omega$ ($0 \leq \omega \leq 1$) is used to represent the relation between $\mathbf{y}_n$ and the extra uniform distribution. Therefore, equation (1) can be rewritten as

$$p(\mathbf{y}_n) = \frac{\omega}{N} + \frac{(1-\omega)}{M} \sum_{m=1}^{M} g(\mathbf{y}_n; \tilde{\mathbf{x}}_m, \sigma^2 \mathbf{I}). \tag{2}$$

Based on the *i.i.d.* data assumption, the log-likelihood function can be denoted as

$$\ln p(\mathbf{Y}) = \sum_{n=1}^{N} \ln \left[ \frac{\omega}{N} + \frac{(1-\omega)}{M} \sum_{m=1}^{M} g(\mathbf{y}_n; \tilde{\mathbf{x}}_m, \sigma^2 \mathbf{I}) \right]. \tag{3}$$

The nonrigid point set registration can be accomplished by maximizing equation (3) using the famous EM (Expectation-Maximum) algorithm, which can be briefly written as follows,

*E-step*: Compute the correspondence matrix $\mathbf{P} = \{p_{mn}\}_{M \times N}$, which can be denoted as

$$p_{mn} = \exp\left(-\frac{\|\mathbf{y}_n - \tilde{\mathbf{x}}_m\|^2}{2\sigma^2}\right) / \left[ \sum_{k=1}^{M} \exp\left(-\frac{\|\mathbf{y}_n - \tilde{\mathbf{x}}_k\|^2}{2\sigma^2}\right) + c \right], \tag{4}$$

where $c = (2\pi\sigma^2)^{D/2} \frac{\omega}{1-\omega} \frac{M}{N}$. Notably, in our paper, we introduce a corresponding constraint, which is expressed as

$$\sum_{n=1}^{N} p_{mn} = 1. \tag{5}$$

Equation (5) means that each model point has to search its corresponding point in the scene point set.

*M-step*: Calculate the spatial transformation by minimizing the following energy function, which can be represented by

$$Q(\mathbf{\theta},\sigma^2) = \frac{1}{2\sigma^2}\sum_{n=1}^{N}\sum_{m=1}^{M} p_{mn}\|\mathbf{y}_n - f(\mathbf{x}_m,\mathbf{\theta})\|^2 + \frac{MD}{2}\ln\sigma^2. \tag{6}$$

In order to smooth the interpolation function to avoid over-arbitrary spatial transformation, a regularization term can be introduced in equation (6) according to the Tikhonov regurgitation theory [37], which is defined as

$$Q(\mathbf{\theta},\sigma^2) = \frac{1}{2\sigma^2}\sum_{n=1}^{N}\sum_{m=1}^{M} p_{mn}\|\mathbf{y}_n - f(\mathbf{x}_m,\mathbf{\theta})\|^2 + \frac{MD}{2}\ln\sigma^2 + \frac{\lambda}{2}\varphi(f) \tag{7}$$

Through iteratively running E-step and M-step, equation (3) can be gradually maximized. The final results can be employed to recover the spatial transformation.

## III. Fast Optimization of M-step

Usually, it needs $O(MN)$ operations to calculate the correspondences in E-step, and fast Gaussian transform (FGT) can further reduce the computational complexity to $O(M+N)$. The computational cost for correspondence is acceptable in practice. Besides, it can be further accelerated by parallel computation on GPU devices. Therefore, we focus on how to reduce the computational burden on calculating the spatial transformation in M-step. Herein, by preserving the terms that are only related to the transformation function $f$, we rewrite equation (7) as

$$Q_f(\mathbf{\theta}) = \frac{1}{2\sigma^2}\sum_{n=1}^{N}\sum_{m=1}^{M} p_{mn}\|\mathbf{y}_n - f(\mathbf{x}_m,\mathbf{\theta})\|^2 + \frac{\lambda}{2}\varphi(f). \tag{8}$$

The new location is defined as the sum between the initial location and a displacement function $v$ based on Motion Coherent Theory (MCT) [38] [39], which is denoted as

$$f(\mathbf{X}) = \mathbf{X} + v(\mathbf{X}). \tag{9}$$

The regularization of the displacement function can be modeled in the Reproducing Kernel Hilbert Space (RKHS). By using calculus of variation, the optimal displacement function with respect to equation (8) can be defined as the linear combinations of Gaussian kernel functions, which is denoted as

$$v(\mathbf{x}) = \sum_{m=1}^{M} \mathbf{w}_m \phi(\mathbf{x},\mathbf{x}_m), \tag{10}$$

where $\phi(\mathbf{x},\mathbf{x}_m) = \exp(-\|\mathbf{x}-\mathbf{x}_m\|^2/2\beta^2)\mathbf{I}_{D\times D}$, and $\mathbf{I}_{D\times D}$ is an $D\times D$ identity matrix. Equation (8) can be rewritten as follows

$$Q_f(\mathbf{W}) = \frac{1}{2\sigma^2}\sum_{n=1}^{N}\sum_{m=1}^{M} p_{mn}\|\mathbf{y}_n - (\mathbf{x}_m + \phi_m\mathbf{W})\|^2 + \frac{\lambda}{2}\operatorname{tr}(\mathbf{W}^T\mathbf{\Phi}\mathbf{W}), \tag{11}$$

where $\Phi=\left(\exp\left(-\|\mathbf{x}_i-\mathbf{x}_j\|^2/2\beta^2\right)\right)\in\mathbb{R}^{M\times M}$ is a symmetric Gram matrix, and $\phi_m\in\mathbb{R}^{1\times M}$ is the $m-th$ row of $\Phi$. Letting $\dfrac{\partial Q_f(\mathbf{W})}{\partial \mathbf{W}^T}=0$ yield

$$-\frac{1}{\sigma^2}\sum_{m=1}^{M}\phi_m^T\sum_{n=1}^{N}p_{mn}\mathbf{y}_n+\frac{1}{\sigma^2}\sum_{m=1}^{M}\phi_m^T\mathbf{x}_m\sum_{n=1}^{N}p_{mn}+\frac{1}{\sigma^2}\sum_{m=1}^{M}\phi_m^T\phi_m\mathbf{W}\sum_{n=1}^{N}p_{mn}+\lambda\Phi\mathbf{W}=\mathbf{0}. \quad (12)$$

According to the corresponding constraint $\sum_{n=1}^{N}p_{mn}=1$, it can deduce

$$-\sum_{m=1}^{M}\phi_m^T\tilde{\mathbf{y}}_n+\sum_{m=1}^{M}\phi_m^T\mathbf{x}_m+\sum_{m=1}^{M}\phi_m^T\phi_m\mathbf{W}+\lambda\sigma^2\Phi\mathbf{W}=\mathbf{0}, \quad (13)$$

where $\tilde{\mathbf{y}}_m=\sum_{n=1}^{N}p_{mn}\mathbf{y}_n$ can be seen as the newly estimated position that is corresponding to $\mathbf{x}_m$ using scene point sets. The compact form of equation (13) can be denoted as

$$\Phi^T\Phi\mathbf{W}+\lambda\sigma^2\Phi\mathbf{W}=\Phi^T\tilde{\mathbf{Y}}-\Phi^T\mathbf{X}, \quad (14)$$

where $\tilde{\mathbf{Y}}$ is the data matrix that are formed by $\{\tilde{\mathbf{y}}_n\}$. Because $\Phi$ is symmetric, we have $\Phi^T=\Phi$. Therefore, the equation (14) can be simplified as

$$\left(\Phi+\lambda\sigma^2\mathbf{I}_{M\times M}\right)\mathbf{W}=\tilde{\mathbf{Y}}-\mathbf{X}, \quad (15)$$

where $\mathbf{I}_{M\times M}$ is an $M\times M$ identity matrix. By taking eigenvalue decomposition of $\Phi=\mathbf{U}\Lambda\mathbf{U}^T$, it can be obtained

$$\mathbf{U}\left(\Lambda+\lambda\sigma^2\mathbf{I}_{M\times M}\right)\mathbf{U}^T\mathbf{W}=\tilde{\mathbf{Y}}-\mathbf{X}. \quad (16)$$

Because $\mathbf{U}^T=\mathbf{U}^{-1}$, the solution of $\mathbf{W}$ can be denoted as

$$\mathbf{W}=\mathbf{U}\left(\Lambda+\lambda\sigma^2\mathbf{I}_{M\times M}\right)^{-1}\mathbf{U}^T\left(\tilde{\mathbf{Y}}-\mathbf{X}\right). \quad (17)$$

Notably, using equation (17) to calculate $\mathbf{W}$, we only need to update a diagonal matrix $\left(\Lambda+\lambda\sigma^2\mathbf{I}_{M\times M}\right)^{-1}$ with linear computational complexity, and to take matrix multiplication operations. It totally needs $O(2DM^2+DM+M)$ operations. Because $D\ll M$, our method can significantly improve the computational efficiency of nonrigid point registration. Besides, the $\sigma^2$ can be calculated as

$$\sigma^2=\frac{1}{DM}\sum_{n=1}^{N}\sum_{m=1}^{M}\|\mathbf{y}_n-f(\mathbf{x}_m,\mathbf{W})\|^2=\frac{1}{DM}\left(\mathrm{tr}\left(\mathbf{Y}^T\mathrm{d}\left(\mathbf{P}^T\mathbf{1}\right)\mathbf{Y}\right)-2\mathrm{tr}\left(\tilde{\mathbf{Y}}^T\tilde{\mathbf{X}}\right)+\mathrm{tr}\left(\tilde{\mathbf{X}}^T\tilde{\mathbf{X}}\right)\right), \quad (18)$$

where $\tilde{\mathbf{X}}=f(\mathbf{X},\mathbf{W})$ is the transformed model point set and $\mathrm{d}^{-1}(\cdot)$ denotes the inverse diagonal matrix.

*Low-rank approximation*: Equation (17) can be further accelerated by low-rank approximation. By preserving the principal components of matrix $\mathbf{\Phi}$, the approximation reconstruction of $\mathbf{\Phi}$ can be denoted as

$$\tilde{\mathbf{\Phi}} = \tilde{\mathbf{U}}\tilde{\mathbf{\Lambda}}\tilde{\mathbf{U}}^T, \tag{19}$$

$\tilde{\mathbf{\Lambda}} \in \mathbb{R}^{K \times K}$ is composed by the $K$ largest eigenvalues of the matrix $\mathbf{\Phi}$, and $\tilde{\mathbf{U}} \in \mathbb{R}^{M \times K}$ is formed by corresponding eigenvectors. Therefore, the approximation solution $\tilde{\mathbf{W}}$ can be denoted as

$$\tilde{\mathbf{W}} = \tilde{\mathbf{U}}\left(\tilde{\mathbf{\Lambda}} + \lambda\sigma^2 \mathbf{I}_{K \times K}\right)^{-1} \tilde{\mathbf{U}}^T \left(\tilde{\mathbf{Y}} - \mathbf{X}\right). \tag{20}$$

The computational complexity of equation (20) is $O(2DKM + DK + K)$ operations. If there are an amount of feature points and the shape of the point cloud is well persevered, we can choose $K \ll M$, which can further reduce the computational burden to be approximately linear.

*Relationship with CPD*: Our method can be simply seen as a fast implementation of CPD so that we have to claim that our method is very likely to CPD, including the theory, cost function and optimization scheme. The main difference is on calculating the spatial transformation. In CPD, the spatial transformation is calculated by

$$\mathbf{W}_{cpd} = \left(\mathbf{\Phi} + \lambda\sigma^2 \mathrm{d}(\mathbf{P1})^{-1}\right)^{-1} \left(\mathrm{d}(\mathbf{P1})^{-1}\mathbf{PY} - \mathbf{X}\right). \tag{21}$$

As shown in equation (21), CPD has to calculate the inversion of a $M \times M$ matrix with time complexity $O(M^3)$. Through introducing the corresponding constraint, our method can get the solutions by updating a diagonal matrix with linear computational complexity, and taking matrix multiplication operation. The matrix-inverse operation is avoided so that the computational burden can be significantly reduced.

**IV. Experiments**

Our method is implemented in MATLAB, and the experimental environment is an Intel Core i7-8750H CPU and 32GB RAM. Our method can be simply treated as a fast implementation of CPD. Besides, the CPD-based methods can be easily accelerated by our method with minor modification. Therefore, the experiments focus on comparing the time cost and the accuracy between CPD and ours. Besides, we also test another recently CPD-based point set registration algorithm that has publicly available codes that are provided by the authors, which is GLTP [Song, 2019]. On the scheme of CPD, GLTP employs the LLE technique to preserve local structures.

*Evaluation Criterion*: We use two experiments to verify the performance of the algorithms. In the first experiment, we adopt the wolf dataset to compare the abilities of the algorithms when confronted with various data degradations, including nonrigid deformation, noise, occlusion and outliers. Besides, we also compare the runtime of the methods. The second experiment focus on comparing the computational efficiency between CPD, CPD(low-rank), fast-CPD and fast-CPD(low-rank) using the bunny dataset. For convenience, we denote CPD(low-rank) and fast-CPD(low-rank) as CPD-L and fast-CPD-L, respectively.

1) Accuracy: The root-mean-square error (RMSE) is used as the registration error, which is denoted as follows,

$$RMSE = \sqrt{\frac{1}{M}\sum_{m=1}^{M}\left\|f(\mathbf{x}_m) - \hat{\mathbf{y}}_m\right\|^2}, \quad (22)$$

where $\hat{\mathbf{y}}_m$ is the ground truth corresponding point of $\mathbf{x}_m$.

2) Runtime: The total runtime $t_{total}$ of the methods can be roughly divided into three parts: the time $t_c$ to calculate the correspondences, the time $t_f$ to compute the spatial transformation, and the time $t_o$ for other operations in the programs, and $t_{total} = t_c + t_f + t_o$. For fast-CPD, CPD-L, and fast-CPD-L, $t_f$ is composed by the time $t_{eig}$ to take eigenvalue decomposition of the $\mathbf{\Phi} \in \mathbb{R}^{M \times M}$ before the iteration begins, and the time $t_{iter}$ to compute the spatial transformation in the iterations, and $t_f = t_{eig} + t_{iter}$. Specially, for CPD and GLTP, there are $t_{eig} = 0$, and $t_f = t_{iter}$.

***Note***: Compared with naive CPD, fast-CPD only additionally performs a normalization operation as equation (5) with linear computational complexity for computing the correspondences. Besides, the cost time for correspondence determination can be significantly reduced by FGT (Fast Gaussian Transform) and parallel computing on GPU devices. Because it does not belong to the research scope of this paper, we do not specially accelerate the step of correspondence determination. The main difference between fast-CPD and CPD is how to compute the spatial transformation. Therefore, we take $t_f$ as the key metric to evaluate the computational efficiency of various registration methods.

*Parameter settings*: The free parameters of CPD and fast-CPD are same, including the noise and outlier parameter $\omega$ $(0 \leq \omega \leq 1)$, the parameter $\beta$ that is used to smooth the Gaussian kernels, and the parameter $\lambda$ that is used to control the roles of the regularization term, as illustrated as equation (11). For a fair comparison, we adopt same settings of these parameters and remain unchanged in the following experiments, which are $\omega = 0.7$, $\beta = 2$ and $\lambda = 10$. For GLTP, we use the authors' recommended parameter settings, and also keep theses settings unchanged in the experiments.

## A. Results on wolf dataset

In this experiment, we use the wolf shapes of the TOSCA 3D point data in [40] to compare the performance of CPD, GLTP and the proposed algorithm on accuracy and efficiency. The wolf dataset contains 3 shapes ('wolf0', 'wolf1', and 'wolf2') in different poses, and each shape has 4344 points. The wolf shapes are renormalized into [-1,1]. We employ wolf0 as the model point set, wolf1 and wolf2 as the scene point sets. Wolf0 and wolf1, wolf0 and wolf2 are defined as Group I and Group II, respectively. We introduce four kinds of data degradations, including nonrigid deformation, noise, occlusion and outliers, to evaluate the performance of the registration methods on accuracy and efficiency. All the methods run 100 iterations.

Fig. 1(a) and (b) show the registration results of fast-CPD between wolf0 and wolf1 (Group I), wolf0 and wolf2 (Group II). Notably, the attitude change between wolf0 and wolf1 is large. We can see that fast-CPD can both well align the two groups of point sets. Next, we add Gaussian white noise to wolf1 and wolf2 to make the points randomly deviate their original positions. The mean of the noise is zero and the standard deviation is 0.1. As shown in Fig.1 (c) and (d), fast-CPD still can well recover the spatial transformation between the point sets, although there is noise pollution in the data. As demonstrated in [15], there are two cases in the degradation of occlusion, which are missing points on one side and on both sides. According to equation (5), the corresponding constraint that is introduced for fast-CPD is to force each model point to search its corresponding point in the other point set. When there are missing points in the model point set, or there are few missing points in the scene point set, our method can keep good performance. When there are a number of missing points in the scene point set, we can first recover the transformation from the scene point set to the model point set to get accurate one-to-one correspondences, and pick out the corresponding points. Then we can utilize the obtained one-to-one correspondences and the selected points to calculate the spatial transformation from the model point set to scene point set. Therefore, our method can well handle the first case. However, our method performs not well when there are many missing points on both sides. In the test of occlusion, we present the experiments with missing points on one side. The number of missing points in wolf0 is 1000. As shown in Fig. (e) and (f), we can see that our method can accurately align the model point sets to the scene point sets. In the test of outliers, we add outliers to wolf1 and wolf2, and the ratio of outlier to data is 0.6. The registration results are shown in Fig. (g) and (h). Our method can perform very well under the interference of outliers. Further, we give the quantitative results of CPD, GLTP and ours in TABLE I. As illustrated in TABLE I, in the test of deformation, our method is slightly better than CPD and GLTP. In the test of noise, the registration results of the three methods are almost the same. In the test of occlusion, we can see that CPD gets the best performance in the experiment of Group II, but fails in Group I. GLTP and ours can keep stable performance in both two experiments, and GLTP is slightly more accurate than ours. In the test of outliers, we can see that the registration results of our method are much better than CPD and GLTP. This shows that the corresponding constraint of equation (5) can help the registration methods to improve their performance in the scenarios of outlier degradation.

In TABLE II and TABLE III, we give the comparisons of runtime on the wolf dataset. As illustrated in the two tables, the time $t_c$ for correspondence determination is relevant to the number of model points and scene points. Compared with CPD and GLTP, fast-CPD consumes a little more time ($t_c$) on correspondence determination because of the extra operations of normalization as illustrated by equation (5). As illustrated by TABLE II and TABLE III, for fast-CPD, the time $t_f$ for spatial transformation retrieval is composed by the time $t_{eig}$ and the time $t_{iter}$. Before the iteration begins, fast-CPD has to use time $t_{eig}$ to take the eigen-decomposition of the kernel matrix $\Phi \in \mathbb{R}^{M \times M}$ once. Although CPD and GLTP are not necessary to do this, they have to compute the inversion of a $M \times M$ matrix with time complexity $O(M^3)$ in each iteration. Therefore, fast-CPD spends much less time on computing the spatial transformation than CPD and GLTP. By comparing the total time consumption $t_{total}$, fast-CPD can run more efficiently than CPD and GLTP in various data degradations.

*B. Results on bunny dataset*

In this experiment, we concentrate on evaluating the computational efficiency of the point set registration methods, including CPD, CPD(low-rank) that is denoted as CPD-L, fast-CPD and fast-CPD(low-rank) that is denoted as fast-CPD-L. We resample the bunny dataset [41] as six sets with different number of points: $4000 \times 3$, $8000 \times 3$, $12000 \times 3$, $16000 \times 3$, $20000 \times 3$ and $24000 \times 3$. The original bunny point set is employed as the scene point set. We adopt the version of bunny point set with random affine transformation as the model point set. Each method runs 50 iterations. CPD-L and fast-CPD-L both preserve $K = 0.1M$ largest eigenvalues of the matrix $\Phi$. For CPD-L, it needs $O(K^3)$ operations to compute the spatial transformation in each iteration. Fig. 2 shows the input point sets and the visualization of the registration results. All the methods are successful to achieve the registration with error less than $5 \times 10^{-3}$. TBALE IV gives the runtime of the four point set registration methods. In the following, we detailly discuss the results.

1) $t_c$. $t_c$ represents the time consumption that is used to determine the corresponding relationship between the two point sets. As illustrated by the third row of TABLE IV, CPD and CPD-L consume less time on the correspondence determination than fast-CPD and fast-CPD-L because fast-CPD and fast-CPD-L have to take normalization operations as equation (5) with linear computational complexity.

   $t_{eig}$. $t_{eig}$ denotes the time consumption that is used to take the eigen-decomposition of the kernel matrix $\Phi \in \mathbb{R}^{M \times M}$. This operation only needs to be taken once. Besides, it can be precomputed to significantly increase the processing speed when there are many different scene point sets to be matched with several model point sets. In our experiment, the eigen-decomposition of the kernel matrix $\Phi \in \mathbb{R}^{M \times M}$ is calculated in advance. In the experiment, CPD-L, fast-CPD and fast-CPD-L can directly load the corresponding data file so that the three methods have the same $t_{eig}$.

2) $t_{iter}$. $t_{iter}$ is the time that is consumed to solve the interpolation functions in the iterations. As illustrated by TABLE IV, although the time $t_{eig}$ of CPD is zero, CPD has to take much more time in the iterations because it needs to perform inversion of a $M \times M$ matrix per iteration with time complexity $O(M^3)$. CPD-L is apparently faster than CPD. Compared with CPD and CPD-

L, fast-CPD spends much less time to compute the spatial transformation. By introducing the low-rank approximation, fast-CPD-L can further reduce the time $t_{iter}$. Notably, fast-CPD-L only needs about 5.1 seconds to calculate the spatial transformation between the point sets that both contain 24000 points, and each iteration just needs about 0.102 seconds.

3) $t_f$. $t_f$ is the sum of $t_{eig}$ and $t_{iter}$. Although CPD-L, fast-CPD and fast-CPD-L have to take eigenvalue decomposition of $\Phi \in \mathbb{R}^{M \times M}$ once before the iterations, but they save a lot of time in the iterations. Besides, in many scenarios, the eigenvalue decomposition of $\Phi \in \mathbb{R}^{M \times M}$ can be precomputed so that the time of $t_f$ can be further lowered. Additionally, fast-CPD and fast-CPD-L are much faster than CPD-L.

4) $t_{total}$. $t_{total}$ is the total time consumption of the point set registration methods. Because the significantly advantage in the step of spatial transformation recovery, fast-CPD and fast-CPD-L are apparently faster than CPD and CPD-L.

In summary, the proposed methods can significantly accelerate the nonrigid registration process, especially in large-scale problems.

## V. Conclusion

In this paper, we propose a fast implementation of CPD, which is namely fast-CPD. The highlight is that our method can avoid the big matrix-inverse operation in each iteration. This can significantly reduce the computational cost. The computational complexity of fast CPD is approximately $O(M^2)$. We can further reduce the computational cost by low-rank approximation. Besides, the modified methods based on CPD can be easily accelerated by our method. The experimental results on 3D point sets show that our method can achieve comparable performance with CPD on accuracy, while it can significantly save the computing time.

The main limitation of the proposed methods is that when there are many miss points are on both sides, our method cannot keep good performance.

## ACKNOWLEDGMENT

The authors thank Andriy Myronenko, Xubo Song, Ge Song and Fan Guoliang for providing their source codes and datasets, which greatly facilitated the comparison experiments.

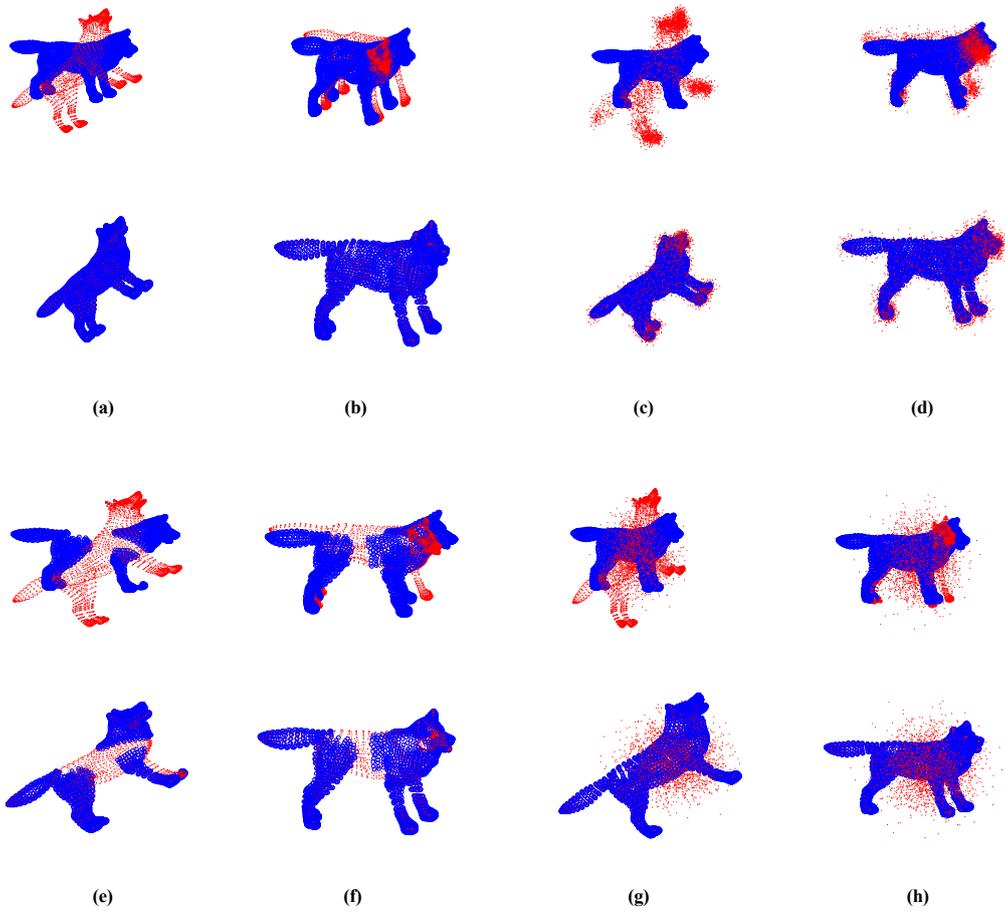

**FIGURE 1.** Registration results of fast-CPD on wolf dataset. (a)-(b): deformation, (c)-(d): noise, (e)-(f): occlusion, (g)-(h): outliers. (a)(c)(e)(f) belong to Group I, and (b)(d)(f)(h) belong to Group II. The upper row shows the data, and the lower row shows the registration results.

TABLE I RMSE of the point set registration methods on wolf dataset.

|  |  | Deformation | Noise | Occlusion | Outliers |
|---|---|---|---|---|---|
| Group I | CPD | 0.0101 | 0.0720 | 0.2239 | 0.0383 |
|  | GLTP | 0.0099 | **0.0715** | **0.0114** | 0.0445 |
|  | Fast-CPD | **0.0087** | 0.0721 | 0.0140 | **0.0090** |
| Group II | CPD | 0.0133 | 0.0715 | **0.0089** | 0.0441 |
|  | GLTP | 0.0113 | **0.0711** | 0.0100 | 0.0496 |
|  | Fast-CPD | **0.0097** | 0.0715 | 0.0102 | **0.0103** |

TABLE II Runtime (s) of the point set registration methods on wolf dataset of Group I.

|  |  | $t_c$ | $t_{eig}$ | $t_{iter}$ | $t_f$ | $t_{total}$ |
|---|---|---|---|---|---|---|
| Deformation | CPD | **35.141** | 0 | 64.706 | 64.706 | 102.636 |
|  | GLTP | 35.445 | 0 | 86.706 | 86.706 | 124.654 |
|  | Fast-CPD | 37.097 | 5.808 | **1.733** | **7.541** | **47.037** |
| Noise | CPD | 38.305 | 0 | 63.725 | 63.725 | 63.725 |
|  | GLTP | **38.205** | 0 | 87.395 | 87.395 | 127.608 |
|  | Fast-CPD | 40.103 | 5.830 | **1.747** | **7.577** | **50.182** |
| Occlusion | CPD | 29.143 | 0 | 34.880 | 34.880 | 65.773 |
|  | GLTP | **28.103** | 0 | 48.316 | 48.316 | 78.047 |
|  | Fast-CPD | 29.207 | 2.518 | **1.062** | **3.58** | **34.396** |
| Outliers | CPD | **59.682** | 0 | 63.399 | 63.399 | 125.761 |
|  | GLTP | 59.977 | 0 | 86.705 | 86.705 | 148.885 |
|  | Fast-CPD | 61.979 | 5.828 | **1.776** | **7.604** | **72.120** |

TABLE III Runtime (s) of the point set registration methods on wolf dataset of Group II.

|  |  | $t_c$ | $t_{eig}$ | $t_{iter}$ | $t_f$ | $t_{total}$ |
|---|---|---|---|---|---|---|
| Deformation | CPD | **34.895** | 0 | 64.417 | 64.417 | 101.954 |
|  | GLTP | 35.850 | 0 | 88.942 | 88.942 | 127.039 |
|  | Fast-CPD | 36.819 | 5.805 | **1.742** | **7.547** | **46.881** |
| Noise | CPD | **37.275** | 0 | 62.573 | 62.573 | 102.418 |
|  | GLTP | 37.416 | 0 | 87.225 | 87.225 | 126.510 |
|  | Fast-CPD | 39.336 | 5.735 | **1.755** | **7.49** | **49.335** |
| Occlusion | CPD | **27.275** | 0 | 34.600 | 34.600 | 63.636 |
|  | GLTP | 27.791 | 0 | 48.476 | 48.476 | 77.911 |
|  | Fast-CPD | 28.362 | 2.589 | **1.042** | **3.631** | **33.571** |
| Outliers | CPD | **59.842** | 0 | 63.927 | 63.927 | 127.476 |
|  | GLTP | 60.117 | 0 | 87.591 | 87.591 | 149.912 |
|  | Fast-CPD | 61.788 | 5.733 | **1.785** | **7.518** | **71.826** |

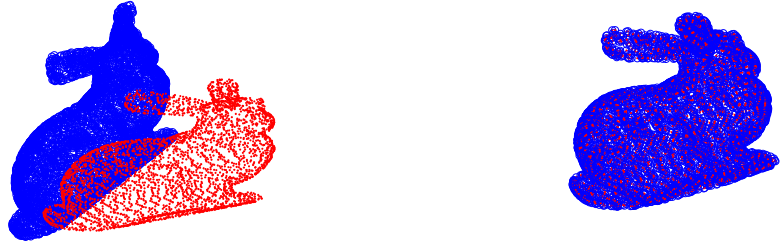

FIGURE 2. Registration results of fast-CPD on bunny dataset with 4000 points. Left: the origin data. Right: the registration results.

TABLE IV Runtime (s) of the point set registration methods on bunny dataset.

| Number of points | | $t_c$ | $t_{eig}$ | $t_{iter}$ | $t_f$ | $t_{total}$ |
|---|---|---|---|---|---|---|
| 4000 | CPD | 15.257 | 0 | 27.529 | 27.529 | 44.161 |
|  | CPD-L | **14.754** | 4.580 | 6.226 | 10.806 | 26.570 |
|  | Fast-CPD | 16.105 | 4.580 | 0.809 | 5.389 | 23.157 |
|  | Fast-CPD-L | 16.057 | 4.580 | **0.106** | **4.686** | **22.359** |
| 8000 | CPD | **61.283** | 0 | 161.938 | 161.938 | 228.066 |
|  | CPD-L | 61.375 | 34.301 | 43.395 | 77.696 | 141.496 |
|  | Fast-CPD | 65.920 | 34.301 | 3.297 | 37.598 | 108.800 |
|  | Fast-CPD-L | 66.143 | 34.301 | **0.374** | **34.675** | **105.812** |
| 12000 | CPD | **136.828** | 0 | 480.412 | 480.412 | 628.192 |
|  | CPD-L | 137.258 | 113.351 | 125.135 | 238.486 | 380.742 |
|  | Fast-CPD | 150.717 | 113.351 | 8.734 | 121.725 | 284.480 |
|  | Fast-CPD-L | 148.652 | 113.351 | **0.766** | **114.117** | **273.534** |
| 16000 | CPD | **240.522** | 0 | 1090.468 | 1090.468 | 1353.219 |
|  | CPD-L | 243.605 | 256.055 | 279.499 | 553.554 | 787.932 |
|  | Fast-CPD | 262.316 | 256.055 | 17.887 | 273.942 | 557.880 |
|  | Fast-CPD-L | 265.928 | 256.055 | **1.713** | **257.768** | **543.163** |
| 20000 | CPD | **373.064** | 0 | 2045.739 | 2045.739 | 2457.879 |
|  | CPD-L | 373.195 | 502.663 | 523.244 | 1025.907 | 1416.212 |
|  | Fast-CPD | 418.746 | 502.663 | 34.592 | 537.255 | 988.430 |
|  | Fast-CPD-L | 415.051 | 502.663 | **3.540** | **506.203** | **956.779** |
| 24000 | CPD | **551.761** | 0 | 3474.422 | 3474.422 | 4119.303 |
|  | CPD-L | 553.102 | 852.925 | 905.934 | 1758.295 | 2364.277 |
|  | Fast-CPD | 630.526 | 852.925 | 47.302 | 900.227 | 1602.056 |
|  | Fast-CPD-L | 615.869 | 852.925 | **5.131** | **858.056** | **1545.760** |